\documentclass[11pt]{article}
\usepackage[preprint]{acl}

\usepackage{times}
\usepackage{latexsym}
\usepackage{algorithm}
\usepackage{algorithmic}
\usepackage{amsmath} 
\usepackage{xcolor}
\usepackage[T1]{fontenc}

\usepackage[utf8]{inputenc}
\usepackage{microtype}

\usepackage{inconsolata}

\usepackage{graphicx}

\title{Does Traversal Order Matter? A Systematic Study of Tree Traversal Methods in Transformer Grammars}

\author{Zongru Liu\thanks{\ \ Equal contribution.}, Pengyu Ji\footnotemark[1], Pengcheng Wang, Kewei Tu\thanks{\ \ Corresponding author} \\
  School of Information Science and Technology, ShanghaiTech University \\
  \texttt{\{liuzr2023,jipy2023,wangpch2024,tukw\}@shanghaitech.edu.cn} \\}

\begin{document}
\maketitle
\begin{abstract}
Transformer Grammars (TGs) enhance language modeling by incorporating syntactic tree structures. Despite the potentially significant impact on model performance of how syntactic trees are linearized in TGs, existing studies rely solely on Depth-First Traversal (DFT) for linearization. In this paper, we expand the traversal design space by exploring Breadth-First Traversal (BFT) and a novel hybrid traversal strategy, Production-Rule Traversal (PRT), which combines the structural lookahead of BFT with the early lexical generation of DFT. We integrate these traversal methods with varying tree configurations and masking strategies, and empirically evaluate their performance on language modeling, syntactic generalization and summarization. We reveal the inherent trade-offs between nested composition and global lookahead, providing actionable recommendations for designing task-aware Transformer Grammars.
\end{abstract}

\section{Introduction}

The transformer architecture \citep{Vaswani-etal-2017-attention} has shown strong performance but lacks the inductive bias of syntactic structure, which is critical for effective generalization \citep{everaert-etal-2015-structures}. Syntactic language models (SLMs) \citep{qian-etal-2021-structural,sartran-etal-2022-transformer, zhao-etal-2024-dependency,hu-etal-2024-generative,ji-etal-2025-tree} address this by introducing syntactic structural inductive biases, modeling linearized parse trees alongside surface sentences.
Crucially, the chosen traversal method dictates the syntactic context, fundamentally shaping the model's syntactic bias. 
However, the traversal method used by existing SLMs to linearize trees into sequences remains underexplored beyond Depth-First Traversal (DFT) \citep{zhao-etal-2025-systematic}.


We systematically explore this traversal space within Transformer Grammars (TGs). Specifically, we investigate DFT and Breadth-First Traversal (BFT). Both approaches exhibit contrasting limitations tied to tree structure. DFT favors early lexical generation but lacks structural lookahead—the ability to access global syntactic information before generating lexical tokens—forcing premature lexical predictions in flat n-ary trees. Conversely, BFT favors early syntactic generation but delays terminal generation, forcing structural predictions without concrete words in deep binary trees. To balance these extremes, we propose Production-Rule Traversal (PRT), fusing BFT's structural lookahead with DFT's early lexical generation.

We evaluate these three traversal methods across varying tree configurations and masking strategies.
The experimental results reveal that the optimal traversal strategy is task-dependent.
Specifically, downstream generation tasks benefit significantly from the structural lookahead of PRT, which provides a broader context for next-token prediction. In contrast, performance on syntactic generalization indicates that syntax-focused tasks favor the strictly nested composition of DFT. 
\section{Related Work}

\textbf{Compositional SLMs.} Transformer Grammars \citep{sartran-etal-2022-transformer, hu-etal-2024-generative} inject structural biases by explicitly modeling and composing trees alongside text. \citet{zhao-etal-2025-systematic} recently unified this paradigm under the Compositional SLM framework. While their study explores top-down and bottom-up linearizations, both remain confined to DFT. DFT lacks structural lookahead, leaving the balance between global syntactic information and early lexical generation unexplored in SLMs.

\textbf{Traversal Strategies.} Traversal strategies profoundly impact structured prediction, from exploiting hierarchical parallelism in non-autoregressive generation \citep{ji-etal-2025-tree} to balancing lookahead and compositionality in discriminative parsing \citep{liu-zhang-2017-order}. We extend these insights to generative SLMs, systematically analyzing how traversal orders affect downstream performance.

\section{Background}
A Transformer Grammar (TG) jointly models a surface sentence $x$ and its constituency tree $y$. For standard autoregressive processing, the tree is linearized to sequences of structural and lexical actions $a = (a_0, \dots, a_{L-1})$ of length $L$, factorizing the joint probability as $p(x, y) = \prod_{i} p(a_i \mid a_{<i})$.

\subsection{Parse Tree Binarization}
While natural constituency trees are inherently non-binary, prior syntactic language models often adopt binarized structures for practical efficiency. Following the established paradigm of \citet{zhao-etal-2025-systematic}, tree binarization is typically evaluated under two configurations: (i) \textbf{Non-binary (N)} trees, which eliminate unary chains directly connected to terminals, so structures ``(NP cat NP)'' are simplified to ``cat'', and (ii) \textbf{Binary (B)} trees, constructed via standard left-binarization.

\subsection{Linearization}

Previous works predominantly rely on top-down Depth-First Traversal (DFT) to linearize syntactic trees \citep{dyer-etal-2016-recurrent}.
Under this paradigm, the action space consists of three types: (i) opening a non-terminal, (ii) generating a terminal token, and (iii) closing a non-terminal. DFT recursively completes all actions for a subtree before executing any actions for its siblings.

\subsection{Composition and Masking}
A prevalent technique for composition in TGs is the internal composition mechanism \citep{sartran-etal-2022-transformer}. In this framework, a closing non-terminal ``X)'' triggers a COMPOSE operation: the model's attention is restricted to the non-terminal's immediate subconstituents, encoding them into a single representation of the subtree. To resume generation, a duplicate ``X)'' immediately follows, attending to the broader context to allow the model to continue predicting subsequent tokens.

Following composition, models adopt one of two masking strategies for subsequent steps: (i) \textbf{Masked (M):} Blocks attention to the internal nodes of composed subconstituents \citep{sartran-etal-2022-transformer}. (ii) \textbf{Not masked (NM):} Retains full attention to all prior tokens \citep{hu-etal-2024-generative}.

\subsection{Inference}
To mitigate the probability imbalance between high-entropy terminals and low-entropy non-terminals, decoding processes often rely on word-synchronous beam search \citep{stern-etal-2017-effective}. Additionally, structural constraint hyperparameters (e.g., maximum non-terminal count and maximum consecutive opening parentheses) identified by \citet{zhao-etal-2025-systematic} are typically applied to prevent the model from over-generating structural actions.

\section{Design Space of Traversal Strategies}

Having described the foundational mechanics of Transformer Grammars, we now outline the design space for evaluating traversal strategies. To assess the impact of explicit syntactic features, we evaluate both \textbf{labeled trees (L)}, which retain the original non-terminals \citep{sartran-etal-2022-transformer}, and \textbf{unlabeled trees (UL)}, which use anonymized labels \citep{zhao-etal-2025-systematic}.

\subsection{Linearization}
\begin{figure}
    \centering
    \includegraphics[width=0.9\linewidth]{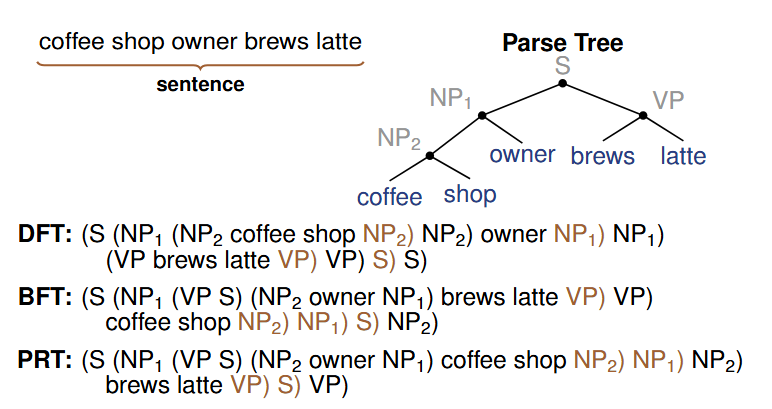}
    \caption{A comparison of tree traversal strategies to linearize an example sentence. We use brown for the closing non-terminal to trigger the COMPOSE operation.}
    \label{fig:traversal_example}
    \vspace{-5pt}
\end{figure}

The chosen traversal order dictates the sequence of linearization actions, strictly determining the syntactic context at any given generation step. To expand the design space beyond the standard depth-first baseline, we introduce two alternative linearization strategies guided by different traversals (exemplified in Figure \ref{fig:traversal_example} and formalized in Appendix \ref{sec:appendix_algorithms}):


\textbf{BFT:}
Under BFT, the linearization operates level by level. Specifically, upon visiting a non-terminal node, the opening actions or terminal generations for all its immediate children are executed sequentially, immediately followed by the closing action for the current node.

\textbf{PRT:}
Under PRT, linearization begins by pushing the root node onto a stack.
At each step, a node is popped, and the opening actions or terminal generations for all its immediate children are executed sequentially. Subsequently, the non-terminal children are pushed in reverse order, immediately followed by the closing action for the current node.

\begin{figure}[t]
\centering
    \includegraphics[width=\columnwidth]{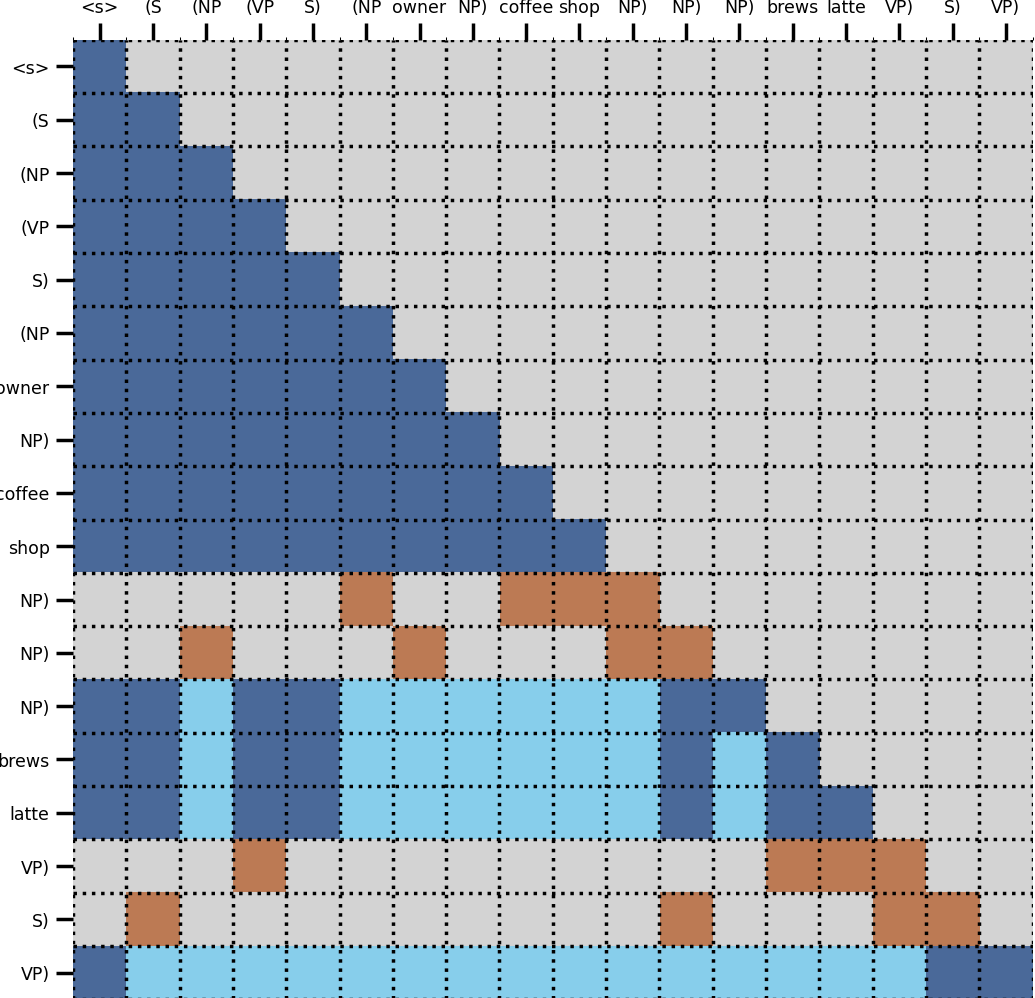} 
    \caption{Illustration of the PRT compose method, which is shared with BFT. We use brown for the attention ranges of internal compositions, dark blue for ordinary attended positions, light blue for already-composed positions that are only accessible in NM.}
    \label{fig:traversal_methods}
    \vspace{-5pt}
\end{figure}
\subsection{Composition}
\label{sec:BFT_compose}
In BFT and PRT, operations for closing non-terminals are strictly dictated by the completion status of their corresponding subtrees. If a subtree remains incomplete (i.e., it contains unclosed non-terminals), the closing non-terminal is not duplicated; instead, it is used for next-token prediction, delaying composition. Conversely, once nested subtrees are fully completed, this delay is resolved by introducing a sequence of duplicated closing non-terminals corresponding to each completed subtree. These execute COMPOSE operations from bottom to top, followed by the current non-terminal resuming next-token prediction, shown in Figure \ref{fig:traversal_methods}.
\begin{table}[t]
  \centering
  \resizebox{0.9\columnwidth}{!}{
  \begin{tabular}{l cccc}
    \hline
    \textbf{Model Variant} & \textbf{B-L} & \textbf{B-UL} & \textbf{N-L} & \textbf{N-UL} \\
        \hline
    \quad DFT-M    & 34.24 & 23.32 & 24.22 & 23.45 \\
    \quad DFT-NM   & \textbf{30.17} & 20.57 & 22.06 & \textbf{21.74} \\
    \quad DFT-tree & 30.32 & 21.40 & 22.40 & 22.20 \\
    \hline
    \quad BFT-M    & 36.88 & 24.45 & 24.57 & 24.27 \\
    \quad BFT-NM   & 32.00 & 22.11 & 22.82 & 22.76 \\
    \quad BFT-tree & 33.18 & 23.00 & 23.90 & 23.01 \\
    \hline
    \quad PRT-M    & 35.22 & 22.83 & 24.12 & 23.59 \\
    \quad PRT-NM   & 30.79 & \textbf{20.38} & \textbf{21.83} & 21.87 \\
    \quad PRT-tree & 30.96 & 21.47 & 22.93 & 22.38 \\
    \hline
  \end{tabular}
  }
  \caption{\label{tab:ppl_results}
Perplexity ($\downarrow$) results of our models, where all values represent upper bounds.
  }
    \vspace{-5pt}
  
\end{table}
\subsection{Inference}
Lacking a static parse tree during inference, our decoding algorithm adapts word-synchronous beam search to track subtree completion. It monitors the output sequence, detects when nested structures complete, and dynamically appends the required duplicated closing non-terminals to activate the composition mechanism described in Section \ref{sec:BFT_compose}.
\section{Experiments}
\begin{table}[t]
  \centering
  \resizebox{0.8\columnwidth}{!}{
  \begin{tabular}{l ccc}
    \hline
    \textbf{Model Variant} & \textbf{N-L} & \textbf{N-UL} & \textbf{B-UL} \\
    \hline
    \quad DFT-M    & \textbf{71.38} & \textbf{71.25} & \textbf{72.77} \\
    \quad DFT-NM   & 71.19 & 69.19 & 70.82 \\
    \quad DFT-tree & 70.38 & 69.37 & 69.69 \\
    \hline
    \quad BFT-M    & 69.86 & 70.84 & 71.21 \\
    \quad BFT-NM   & 69.82 & 69.64 & 69.23 \\
    \quad BFT-tree & 68.79 & 67.59 & 69.12 \\
    \hline
    \quad PRT-M    & 70.90 & 71.12 & 71.86 \\
    \quad PRT-NM   & 70.09 & 70.11 & 70.48 \\
    \quad PRT-tree & 69.24 & 69.16 & 70.24 \\
    \hline
  \end{tabular}
  }
  \caption{\label{tab:blimp_results}
    BLiMP ($\uparrow$) results of our models
  }
    \vspace{-5pt}
  
\end{table}
\begin{table*}[ht]
  \centering
  \resizebox{0.8\textwidth}{!}{
  \begin{tabular}{l cccc cccc cccc}
    \hline
    & \multicolumn{4}{c}{\textbf{N-L}} & \multicolumn{4}{c}{\textbf{N-UL}} & \multicolumn{4}{c}{\textbf{B-UL}} \\
    \cline{2-5} \cline{6-9} \cline{10-13}
    \textbf{Model Variant} & \textbf{R1} & \textbf{R2} & \textbf{RL} & \textbf{R-AVG} & \textbf{R1} & \textbf{R2} & \textbf{RL} & \textbf{R-AVG} & \textbf{R1} & \textbf{R2} & \textbf{RL} & \textbf{R-AVG} \\
    \hline
    \quad DFT-tree & 30.77 & 10.04 & 24.10 & 21.64 & 30.26 & 10.10 & 24.02 & 21.46 & 29.31 & 9.52 & 23.16 & 20.66 \\
    \quad DFT-M    & 26.61 & 7.85 & 21.22 & 18.56 & 27.38 & 8.36 & 21.88 & 19.21 & 25.26 & 7.09 & 19.99 & 17.44 \\
    \quad DFT-NM   & 30.58 & 10.35 & 24.40 & 21.78 & 30.23 & 10.09 & 24.08 & 21.46 & 29.69 & 9.80 & 23.51 & 21.00 \\
    \hline
    \quad BFT-tree & 30.47 & 10.06 & 24.07 & 21.53 & 30.63 & 9.97 & 23.96 & 21.52 & 29.88 & 9.42 & 23.15 & 20.02 \\
    \quad BFT-M    & 27.46 & 7.98 & 21.56 & 19.00 & 27.28 & 7.92 & 21.44 & 18.88 & 26.18 & 7.15 & 20.31 & 17.88 \\
    \quad BFT-NM   & 31.04 & 10.34 & 24.44 & 21.94 & 30.08 & 10.32 & 24.33 & 21.91 & 30.31 & 9.62 & 23.37 & 21.10 \\
    \hline
    \quad PRT-tree & \textbf{31.31} & \textbf{10.47} & 24.64 & 22.14 & 30.45 & 10.02 & 23.95 & 21.47 & 30.03 & 9.56 & 23.35 & 20.98 \\
    \quad PRT-M    & 27.50 & 8.18 & 21.75 & 19.14 & 27.34 & 7.97 & 21.50 & 18.94 & 26.01 & 7.42 & 20.39 & 17.94 \\
    \quad PRT-NM   & 31.29 & \textbf{10.47} & \textbf{24.71} & \textbf{22.16} & \textbf{31.33} & \textbf{10.54} & \textbf{24.65} & \textbf{22.17} & \textbf{30.65} & \textbf{10.33} & \textbf{23.96} & \textbf{21.65} \\
    \hline
  \end{tabular}
  }
  \caption{\label{tab:xsum_results}
    ROUGE scores on the XSum dataset. Best results in each column are bolded.
  }
    \vspace{-5pt}
  
\end{table*}
We evaluate BFT and PRT against all DFT variants, our baselines. All models are trained from scratch on the BLLIP-LG dataset of \citet{Charniak:00} with training splits from \citet{hu-etal-2020-systematic}, parsed via an off-the-shelf standard CRF constituency parser \citep{zhang-etal-2020-fast-and}, implemented in Supar\footnote{\url{https://github.com/yzhangcs/parser}}. We adopt a modernized GPT-2 Small architecture ($\sim$109M parameters) for all variants, deferring comprehensive configurations to Appendix \ref{sec:appendix_hyperparameters}. Across all traversal methods, the standard model trained on the linearized tree sequence is denoted as X-tree (e.g., DFT-tree). Additionally, a traditional token-level language model baseline is provided in Appendix \ref{sec:appendix_vanilla_results}. Crucially, all traversal strategies operate on the exact same underlying constituency trees, differing solely in their traversal orders. 
We first evaluate all variants on standard language modeling, filtering out underperforming configurations before assessing the remainder on syntactic generalization and downstream summarization.

\subsection{Document-Level Language Modeling}
We evaluate all models on the testing split of BLLIP-LG from \citet{hu-etal-2020-systematic}.

\label{sec:experimental_setup}

Computing the exact string probability $p(x) = \sum_y p(x, y)$ is intractable. Following \citet{sartran-etal-2022-transformer}, we approximate sentence-level $p(x)$ by summing over a proposal set of 300 unlabeled constituency trees sampled without replacement via a CRF parser, establishing a strict lower bound for $p(x)$ and a valid upper bound for perplexity. For document-level modeling, marginalizing trees across historical sentences is computationally prohibitive. Thus, we approximate the context for the $i$-th sentence by greedily selecting the single highest-probability tree for each preceding sentence to serve as a fixed structural prefix, following \citet{sartran-etal-2022-transformer}.

As shown in Table \ref{tab:ppl_results}, unmasked models have lower perplexity, with PRT-NM excelling in N-L and B-UL. The B-L configuration fails because forcing the model to predict labels for numerous nodes introduced by binarization consumes an excessive amount of representational capacity, which is consistent with empirical findings in \citet{sartran-etal-2022-transformer}. Therefore, we exclude B-L from downstream evaluations.
\subsection{Syntactic Generalization}

To measure syntactic generalization, we evaluate our models on BLiMP \citep{warstadt-etal-2020-blimp-benchmark}.

Accuracy is measured by whether a model assigns a higher marginal probability $p(x)$ to the grammatical sentence within each minimal pair, approximated using the identical 300-tree sampling method described in Section \ref{sec:experimental_setup}.

For syntactic generalization, masked models yield the highest accuracy, with the baseline DFT-M achieving the best overall performance in all tree structures. The results are presented in Table \ref{tab:blimp_results}.
\subsection{Summarization}
We evaluate text summarization on the XSum dataset \citep{narayan-etal-2018-dont}.

We truncate parsed source documents to 1,800 tokens, followed by a parsed ``Summary above article in one sentence.'' prompt. Models are finetuned on XSum for 10 epochs (batch 40). We report ROUGE \citep{lin-hovy-2003-automatic} using beam search (size 6) with linearized silver parse trees as input.

As shown in Table~\ref{tab:xsum_results}, PRT generally outperforms the BFT and DFT baselines. Across all traversals, unmasked configurations (-NM, -tree) consistently dominate masked ones (-M), culminating in our proposed PRT-NM achieving the highest overall ROUGE scores.

\subsection{Overall Observations}
Based on our experiments, we highlight key findings for syntactic language modeling:

\textbf{(i)} PRT effectively balances structured syntax with sequential generation. For document-level language modeling and summarization tasks, PRT-NM consistently outperforms traditional DFT baselines.

\textbf{(ii)} While BFT's structural lookahead improves summarization, its excessive delay of terminal tokens hurts language modeling and syntactic generalization. This confirms that layer-wise expansion severely disrupts the left-to-right context, highlighting why PRT’s moderate approach is necessary.

\textbf{(iii)} While B-UL minimizes perplexity via dense non-terminals, its increased sequence length degrades long-distance attention and summarization. Compact N-L structures are optimal, balancing length with semantic guidance. Conversely, B-L fails due to the excessive classification burden of predicting labels for numerous binarized nodes.

\section{Conclusion}

In summary, we extend the Transformer Grammar framework by applying BFT and proposing PRT to enable structural lookahead in generative syntactic modeling. Through a systematic empirical evaluation across diverse tree configurations and attention masking mechanisms, we reveal a fundamental bifurcation in structural inductive biases: generation-focused tasks benefit significantly from the global lookahead of PRT, whereas syntax-focused tasks favor the strictly nested composition of DFT.

\section{Limitations}
While our study provides foundational insights, its primary limitations include the restriction to small-scale architectures ($\sim$109M parameters), the reliance on external syntactic parsers, and an exclusive focus on internal composition mechanisms without evaluating external alternatives.

\bibliography{slimmed}

@string{acl = {Association for Computational Linguistics}}

@string{anth = {https://aclanthology.org/}}

@misc{Charniak:00,
    author  = {Eugene Charniak and Don Blaheta and Niyu Ge and Keith Hall and John Hale and Mark Johnson},
    title   = {Bllip 1987-89 wsj corpus release 1},
    year    = "2000",
    note    = {Linguistic Data Consortium, 36}
}

@inproceedings{dyer-etal-2016-recurrent,title = "Recurrent Neural Network Grammars",author = "Dyer, Chris and Kuncoro, Adhiguna and Ballesteros, Miguel and Smith, Noah A.",editor = "Knight, Kevin and Nenkova, Ani and Rambow, Owen",booktitle = "Proceedings of the 2016 Conference of the North {A}merican Chapter of the Association for Computational Linguistics: Human Language Technologies",month = jun,year = "2016",address = "San Diego, California",publisher = acl,url = anth # {N16-1024/},doi = "10.18653/v1/N16-1024",pages = "199--209"}

@article{everaert-etal-2015-structures,
    author = {Everaert, Martin B.H. and Huybregts, Marinus A.C. and Chomsky, Noam and Berwick, Robert C. and Bolhuis, Johan J.},
    title = {Structures, not strings: Linguistics as part of the cognitive sciences},
    journal = {Trends in Cognitive Sciences},
    volume = {19},
    number = {12},
    pages = {729--743},
    year = {2015},
    publisher = {Elsevier}
}

@inproceedings{hu-etal-2020-systematic,title = "A Systematic Assessment of Syntactic Generalization in Neural Language Models",author = "Hu, Jennifer and Gauthier, Jon and Qian, Peng and Wilcox, Ethan and Levy, Roger",editor = "Jurafsky, Dan and Chai, Joyce and Schluter, Natalie and Tetreault, Joel",booktitle = "Proceedings of the 58th Annual Meeting of the Association for Computational Linguistics",month = jul,year = "2020",address = "Online",publisher = acl,url = anth # {2020.acl-main.158/},doi = "10.18653/v1/2020.acl-main.158",pages = "1725--1744"}

@inproceedings{hu-etal-2024-generative,title = "Generative Pretrained Structured Transformers: Unsupervised Syntactic Language Models at Scale",author = "Hu, Xiang and Ji, Pengyu and Zhu, Qingyang and Wu, Wei and Tu, Kewei",editor = "Ku, Lun-Wei and Martins, Andre and Srikumar, Vivek",booktitle = "Proceedings of the 62nd Annual Meeting of the Association for Computational Linguistics (Volume 1: Long Papers)",month = aug,year = "2024",address = "Bangkok, Thailand",publisher = acl,url = anth # {2024.acl-long.145/},doi = "10.18653/v1/2024.acl-long.145",pages = "2640--2657"}

@inproceedings{ji-etal-2025-tree,title = "Tree-Structured Non-Autoregressive Decoding for Sequence-to-Sequence Text Generation",author = "Ji, Pengyu and Liu, Yufei and Hu, Xiang and Tu, Kewei",editor = "Christodoulopoulos, Christos and Chakraborty, Tanmoy and Rose, Carolyn and Peng, Violet",booktitle = "Findings of the Association for Computational Linguistics: EMNLP 2025",month = nov,year = "2025",address = "Suzhou, China",publisher = acl,url = anth # {2025.findings-emnlp.327/},doi = "10.18653/v1/2025.findings-emnlp.327",pages = "6168--6174",ISBN = "979-8-89176-335-7"}

@inproceedings{lin-hovy-2003-automatic,title = "Automatic Evaluation of Summaries Using N-gram Co-occurrence Statistics",author = "Lin, Chin-Yew and Hovy, Eduard",booktitle = "Proceedings of the 2003 Human Language Technology Conference of the North {A}merican Chapter of the Association for Computational Linguistics",year = "2003",url = anth # {N03-1020/},pages = "150--157"}

@article{liu-zhang-2017-order,title = "In-Order Transition-based Constituent Parsing",author = "Liu, Jiangming and Zhang, Yue",editor = "Lee, Lillian and Johnson, Mark and Toutanova, Kristina",journal = "Transactions of the Association for Computational Linguistics",volume = "5",year = "2017",address = "Cambridge, MA",publisher = "MIT Press",url = anth # {Q17-1029/},doi = "10.1162/tacl_a_00070",pages = "413--424"}

@inproceedings{narayan-etal-2018-dont,title = "Don{'}t Give Me the Details, Just the Summary! Topic-Aware Convolutional Neural Networks for Extreme Summarization",author = "Narayan, Shashi and Cohen, Shay B. and Lapata, Mirella",editor = "Riloff, Ellen and Chiang, David and Hockenmaier, Julia and Tsujii, Jun{'}ichi",booktitle = "Proceedings of the 2018 Conference on Empirical Methods in Natural Language Processing",month = oct # "-" # nov,year = "2018",address = "Brussels, Belgium",publisher = acl,url = anth # {D18-1206/},doi = "10.18653/v1/D18-1206",pages = "1797--1807"}

@inproceedings{qian-etal-2021-structural,title = "Structural Guidance for Transformer Language Models",author = "Qian, Peng and Naseem, Tahira and Levy, Roger and Fernandez Astudillo, Ram{\'o}n",editor = "Zong, Chengqing and Xia, Fei and Li, Wenjie and Navigli, Roberto",booktitle = "Proceedings of the 59th Annual Meeting of the Association for Computational Linguistics and the 11th International Joint Conference on Natural Language Processing (Volume 1: Long Papers)",month = aug,year = "2021",address = "Online",publisher = acl,url = anth # {2021.acl-long.289/},doi = "10.18653/v1/2021.acl-long.289",pages = "3735--3745"}

@article{sartran-etal-2022-transformer,title = "Transformer Grammars: Augmenting Transformer Language Models with Syntactic Inductive Biases at Scale",author = "Sartran, Laurent and Barrett, Samuel and Kuncoro, Adhiguna and Stanojevi{\'c}, Milo{\v{s}} and Blunsom, Phil and Dyer, Chris",editor = "Roark, Brian and Nenkova, Ani",journal = "Transactions of the Association for Computational Linguistics",volume = "10",year = "2022",address = "Cambridge, MA",publisher = "MIT Press",url = anth # {2022.tacl-1.81/},doi = "10.1162/tacl_a_00526",pages = "1423--1439"}

@misc{shazeer-2020-glu,
      title={GLU Variants Improve Transformer}, 
      author={Noam Shazeer},
      year={2020},
      eprint={2002.05202},
      archivePrefix={arXiv},
      primaryClass={cs.LG},
      url={https://arxiv.org/abs/2002.05202}, 
}

@inproceedings{stern-etal-2017-effective,title = "Effective Inference for Generative Neural Parsing",author = "Stern, Mitchell and Fried, Daniel and Klein, Dan",editor = "Palmer, Martha and Hwa, Rebecca and Riedel, Sebastian",booktitle = "Proceedings of the 2017 Conference on Empirical Methods in Natural Language Processing",month = sep,year = "2017",address = "Copenhagen, Denmark",publisher = acl,url = anth # {D17-1178/},doi = "10.18653/v1/D17-1178",pages = "1695--1700"}

@article{su-etal-2023-roformer,
title = {RoFormer: Enhanced transformer with Rotary Position Embedding},
journal = {Neurocomputing},
volume = {568},
pages = {127063},
year = {2024},
issn = {0925-2312},
doi = {https://doi.org/10.1016/j.neucom.2023.127063},
url = {https://www.sciencedirect.com/science/article/pii/S0925231223011864},
author = {Jianlin Su and Murtadha Ahmed and Yu Lu and Shengfeng Pan and Wen Bo and Yunfeng Liu}
}

@article{Vaswani-etal-2017-attention,
  title={Attention is all you need},
  author={Vaswani, Ashish and Shazeer, Noam and Parmar, Niki and Uszkoreit, Jakob and Jones, Llion and Gomez, Aidan N and Kaiser, {\L}ukasz and Polosukhin, Illia},
  journal={Advances in neural information processing systems},
  volume={30},
  year={2017}
}

@article{warstadt-etal-2020-blimp-benchmark,title = "{BL}i{MP}: The Benchmark of Linguistic Minimal Pairs for {E}nglish",author = "Warstadt, Alex and Parrish, Alicia and Liu, Haokun and Mohananey, Anhad and Peng, Wei and Wang, Sheng-Fu and Bowman, Samuel R.",editor = "Johnson, Mark and Roark, Brian and Nenkova, Ani",journal = "Transactions of the Association for Computational Linguistics",volume = "8",year = "2020",address = "Cambridge, MA",publisher = "MIT Press",url = anth # {2020.tacl-1.25/},doi = "10.1162/tacl_a_00321",pages = "377--392"}

@inproceedings{zhang-etal-2020-fast-and,
  title     = {Fast and Accurate Neural CRF Constituency Parsing},
  author    = {Zhang, Yu and Zhou, Houquan and Li, Zhenghua},
  booktitle = {Proceedings of the Twenty-Ninth International Joint Conference on
               Artificial Intelligence, {IJCAI-20}},
  publisher = {International Joint Conferences on Artificial Intelligence Organization},
  editor    = {Christian Bessiere},
  pages     = {4046--4053},
  year      = {2020},
  month     = {7},
  note      = {Main track},
  doi       = {10.24963/ijcai.2020/560},
  url       = {https://doi.org/10.24963/ijcai.2020/560},
}

@misc{zhang-sennrich-2019-root,
      title={Root Mean Square Layer Normalization}, 
      author={Biao Zhang and Rico Sennrich},
      year={2019},
      eprint={1910.07467},
      archivePrefix={arXiv},
      primaryClass={cs.LG},
      url={https://arxiv.org/abs/1910.07467}, 
}

@inproceedings{zhao-etal-2024-dependency,title = "Dependency Transformer Grammars: Integrating Dependency Structures into Transformer Language Models",author = "Zhao, Yida and Lou, Chao and Tu, Kewei",editor = "Ku, Lun-Wei and Martins, Andre and Srikumar, Vivek",booktitle = "Proceedings of the 62nd Annual Meeting of the Association for Computational Linguistics (Volume 1: Long Papers)",month = aug,year = "2024",address = "Bangkok, Thailand",publisher = acl,url = anth # {2024.acl-long.84/},doi = "10.18653/v1/2024.acl-long.84",pages = "1543--1556"}

@inproceedings{zhao-etal-2025-systematic,title = "A Systematic Study of Compositional Syntactic Transformer Language Models",author = "Zhao, Yida and Xve, Hao and Hu, Xiang and Tu, Kewei",editor = "Che, Wanxiang and Nabende, Joyce and Shutova, Ekaterina and Pilehvar, Mohammad Taher",booktitle = "Proceedings of the 63rd Annual Meeting of the Association for Computational Linguistics (Volume 1: Long Papers)",month = jul,year = "2025",address = "Vienna, Austria",publisher = acl,url = anth # {2025.acl-long.350/},doi = "10.18653/v1/2025.acl-long.350",pages = "7070--7083",ISBN = "979-8-89176-251-0"}

\appendix

\section{Traversal Algorithm of BFT and PRT}
\label{sec:appendix_algorithms}
To facilitate reproducibility and provide a rigorous formalization of our proposed traversal strategies, we detail the exact procedures for Breadth-First Traversal (BFT) and Production-Rule Traversal (PRT) in Algorithm \ref{alg:bft} and Algorithm \ref{alg:prt}, respectively. Both algorithms take a standard constituency tree as input and deterministically produce a linearized sequence of structural and lexical tokens.

Crucially, the structural divergence between the two methods is dictated by their underlying data structures. BFT utilizes a standard First-In-First-Out (FIFO) queue to ensure strict level-by-level expansion. In contrast, PRT employs a Last-In-First-Out (LIFO) stack.

\begin{algorithm}[ht]
\caption{Breadth-First Traversal (BFT)}
\label{alg:bft}
\textbf{Input:} $T$ root of the constituency tree\\
\textbf{Output:} $R$ sequence of traversal tokens

\begin{algorithmic}[1]
\STATE $Q \leftarrow []$ \hfill $\triangleright$ Empty queue
\STATE $Q\text{.enqueue}(T)$
\STATE $R \leftarrow [\text{OPEN}(T)]$ \hfill $\triangleright$ e.g., \texttt{(X}
\WHILE{$Q \neq \emptyset$}
    \STATE $N \leftarrow Q\text{.dequeue}()$
    \FOR{\textbf{each} child $C$ \textbf{in} $N\text{.children}$}
        \IF{$\text{type}(C) = \text{NonTerminal}$}
            \STATE $Q\text{.enqueue}(C)$
            \STATE $R\text{.append}(\text{OPEN}(C))$
        \ELSE
            \STATE $R\text{.append}(C)$ \hfill $\triangleright$ Terminal token
        \ENDIF
    \ENDFOR
    \STATE $R\text{.append}(\text{CLOSE}(N))$ \hfill $\triangleright$ e.g., \texttt{X)}
\ENDWHILE
\STATE \textbf{return} $R$
\end{algorithmic}
\end{algorithm}
\begin{algorithm}[ht]
\caption{Production-Rule Traversal (PRT)}
\label{alg:prt}
\textbf{Input:} $T$ root of the constituency tree\\
\textbf{Output:} $R$ sequence of traversal tokens

\begin{algorithmic}[1]
\STATE $S \leftarrow []$ \hfill $\triangleright$ Empty stack
\STATE $S\text{.push}(T)$
\STATE $R \leftarrow [\text{OPEN}(T)]$
\WHILE{$S \neq \emptyset$}
    \STATE $N \leftarrow S\text{.pop}()$
    \FOR{\textbf{each} child $C$ \textbf{in} $N\text{.children}$ (left-to-right)}
        \IF{$\text{type}(C) = \text{NonTerminal}$}
            \STATE $R\text{.append}(\text{OPEN}(C))$
        \ELSE
            \STATE $R\text{.append}(C)$
        \ENDIF
    \ENDFOR
    \FOR{\textbf{each} child $C$ \textbf{in} $N\text{.children}$ (right-to-left)}
        \IF{$\text{type}(C) = \text{NonTerminal}$}
            \STATE $S\text{.push}(C)$ \hfill $\triangleright$ Push for depth-first expansion
        \ENDIF
    \ENDFOR
    \STATE $R\text{.append}(\text{CLOSE}(N))$
\ENDWHILE
\STATE \textbf{return} $R$
\end{algorithmic}
\end{algorithm}

\section{Training and Architectural Details}
\label{sec:appendix_hyperparameters}

Our Transformer Grammar models are implemented based on the modernized GPT-2 Small architecture. We adopt several modernized architectural choices standard in recent large language models, including SwiGLU activations \citep{shazeer-2020-glu}, Rotary Position Embeddings \citep{su-etal-2023-roformer}, and RMSNorm \citep{zhang-sennrich-2019-root} for pre-normalization. Furthermore, we tie the weights of the input embedding and the final output projection layer. To accelerate autoregressive generation during inference, we also implement a Key-Value (KV) cache mechanism.

The models are trained on a single NVIDIA A6000 GPU with mixed precision (\texttt{bfloat16}). We optimize the network using AdamW with a cosine learning rate decay schedule and a linear warmup phase. Detailed hyperparameters for both the architecture and the optimization process are comprehensively listed in Table \ref{tab:hyperparams}.
\begin{table}[t]
\centering
\small
\begin{tabular}{lc}
\hline
\textbf{Hyperparameter} & \textbf{Value} \\
\hline
\multicolumn{2}{c}{\textit{Model Architecture}} \\
Number of layers & $12$ \\
Hidden size (d\textsubscript{model}) & $768$ \\
Attention heads & $12$ \\
FFN intermediate size & $1536$ \\ 
Max sequence length & $2048$ \\
Vocabulary size & $50,322$ \\
Activation function & SwiGLU \\
Normalization & RMSNorm \\
Position embeddings & RoPE \\
Bias (Linear \& Norm) & True \\ 
Weight tying & True \\ 
Total Parameters & $\sim$109M\\
\hline
\multicolumn{2}{c}{\textit{Training \& Optimization}} \\
Optimizer & AdamW \\
Adam $\beta$ & $(0.9, 0.95)$ \\
Adam $\epsilon$ & $1 \times 10^{-8}$ \\
Peak learning rate & $6 \times 10^{-4}$ \\
Minimum learning rate & $1 \times 10^{-5}$ \\
Learning rate schedule & Cosine \\
Warmup steps & $300$ \\
Weight decay & $0.2$ \\
Gradient clipping norm & $1.0$ \\
Global batch size & $256$ \\
Microbatch size per device & $16$ \\
Training epochs & $15$ \\
Precision & \texttt{bfloat16} \\
\hline
\end{tabular}

\caption{Architectural and optimization hyperparameters used for our TG models.}
\label{tab:hyperparams}
\end{table}
\section{Traditional Token Baseline Results}

\label{sec:appendix_vanilla_results}
To contextualize the performance of the Transformer Grammars, we additionally train a standard text-only language model baseline (Token) operating purely on the unparsed text. This model shares identical architectural hyperparameters and training configurations as the main models evaluated in the text. 

To facilitate a direct and comprehensive comparison, we present the full experimental results for all traversal methods and tree configurations alongside this traditional Token baseline. Specifically, Table \ref{tab:appendix_ppl} details the document-level language modeling performance measured by perplexity, while Table \ref{tab:appendix_blimp} reports the zero-shot syntactic generalization accuracy on the BLiMP benchmark. Finally, Table \ref{tab:appendix_xsum} provides the abstractive summarization ROUGE scores on the XSum dataset, where the Token baseline is replicated across all tree configuration columns for straightforward vertical comparison. Across all evaluations, the structural models clearly demonstrate superior performance over the purely sequential text baseline.

\begin{table}[t]
  \centering
  \resizebox{0.9\columnwidth}{!}{
  \begin{tabular}{l cccc}
    \hline
    \textbf{Model Variant} & \textbf{B-L} & \textbf{B-UL} & \textbf{N-L} & \textbf{N-UL} \\
    \hline
    \quad Token & \multicolumn{4}{c}{21.90} \\
    \hline
    \quad DFT-M    & 34.24 & 23.32 & 24.22 & 23.45 \\
    \quad DFT-NM   & \textbf{30.17} & 20.57 & 22.06 & \textbf{21.74} \\
    \quad DFT-tree & 30.32 & 21.40 & 22.40 & 22.20 \\
    \hline
    \quad BFT-M    & 36.88 & 24.45 & 24.57 & 24.27 \\
    \quad BFT-NM   & 32.00 & 22.11 & 22.82 & 22.76 \\
    \quad BFT-tree & 33.18 & 23.00 & 23.90 & 23.01 \\
    \hline
    \quad PRT-M    & 35.22 & 22.83 & 24.12 & 23.59 \\
    \quad PRT-NM   & 30.79 & \textbf{20.38} & \textbf{21.83} & 21.87 \\
    \quad PRT-tree & 30.96 & 21.47 & 22.93 & 22.38 \\
    \hline
  \end{tabular}
  }
  \caption{\label{tab:appendix_ppl}
    Comprehensive perplexity ($\downarrow$) results. The text-only Token baseline is provided as a universal reference.
  }
\end{table}

\begin{table}[t]
  \centering
  \resizebox{0.8\columnwidth}{!}{
  \begin{tabular}{l ccc}
    \hline
    \textbf{Model Variant} & \textbf{N-L} & \textbf{N-UL} & \textbf{B-UL} \\
    \hline
    \quad Token & \multicolumn{3}{c}{69.43} \\
    \hline
    \quad DFT-M    & \textbf{71.38} & \textbf{71.25} & \textbf{72.77} \\
    \quad DFT-NM   & 71.19 & 69.19 & 70.82 \\
    \quad DFT-tree & 70.38 & 69.37 & 69.69 \\
    \hline
    \quad BFT-M    & 69.86 & 70.84 & 71.21 \\
    \quad BFT-NM   & 69.82 & 69.64 & 69.23 \\
    \quad BFT-tree & 68.79 & 67.59 & 69.12 \\
    \hline
    \quad PRT-M    & 70.90 & 71.12 & 71.86 \\
    \quad PRT-NM   & 70.09 & 70.11 & 70.48 \\
    \quad PRT-tree & 69.24 & 69.16 & 70.24 \\
    \hline
  \end{tabular}
  }
  \caption{\label{tab:appendix_blimp}
    Comprehensive BLiMP accuracy ($\uparrow$) results. All structural configurations are compared against the traditional Token baseline.
  }
\end{table}

\begin{table*}[t]
  \centering
  \resizebox{0.95\textwidth}{!}{
  \begin{tabular}{l cccc cccc cccc}
    \hline
    & \multicolumn{4}{c}{\textbf{N-L}} & \multicolumn{4}{c}{\textbf{N-UL}} & \multicolumn{4}{c}{\textbf{B-UL}} \\
    \cline{2-5} \cline{6-9} \cline{10-13}
    \textbf{Model Variant} & \textbf{R1} & \textbf{R2} & \textbf{RL} & \textbf{R-AVG} & \textbf{R1} & \textbf{R2} & \textbf{RL} & \textbf{R-AVG} & \textbf{R1} & \textbf{R2} & \textbf{RL} & \textbf{R-AVG} \\
    \hline
    \quad Token & 26.03 & 7.91 & 20.84 & 18.26 & 26.03 & 7.91 & 20.84 & 18.26 & 26.03 & 7.91 & 20.84 & 18.26 \\
    \hline
    \quad DFT-tree & 30.77 & 10.04 & 24.10 & 21.64 & 30.26 & 10.10 & 24.02 & 21.46 & 29.31 & 9.52 & 23.16 & 20.66 \\
    \quad DFT-M    & 26.61 & 7.85 & 21.22 & 18.56 & 27.38 & 8.36 & 21.88 & 19.21 & 25.26 & 7.09 & 19.99 & 17.44 \\
    \quad DFT-NM   & 30.58 & 10.35 & 24.40 & 21.78 & 30.23 & 10.09 & 24.08 & 21.46 & 29.69 & 9.80 & 23.51 & 21.00 \\
    \hline
    \quad BFT-tree & 30.47 & 10.06 & 24.07 & 21.53 & 30.63 & 9.97 & 23.96 & 21.52 & 29.88 & 9.42 & 23.15 & 20.02 \\
    \quad BFT-M    & 27.46 & 7.98 & 21.56 & 19.00 & 27.28 & 7.92 & 21.44 & 18.88 & 26.18 & 7.15 & 20.31 & 17.88 \\
    \quad BFT-NM   & 31.04 & 10.34 & 24.44 & 21.94 & 30.08 & 10.32 & 24.33 & 21.91 & 30.31 & 9.62 & 23.37 & 21.10 \\
    \hline
    \quad PRT-tree & \textbf{31.31} & \textbf{10.47} & 24.64 & 22.14 & 30.45 & 10.02 & 23.95 & 21.47 & 30.03 & 9.56 & 23.35 & 20.98 \\
    \quad PRT-M    & 27.50 & 8.18 & 21.75 & 19.14 & 27.34 & 7.97 & 21.50 & 18.94 & 26.01 & 7.42 & 20.39 & 17.94 \\
    \quad PRT-NM   & 31.29 & \textbf{10.47} & \textbf{24.71} & \textbf{22.16} & \textbf{31.33} & \textbf{10.54} & \textbf{24.65} & \textbf{22.17} & \textbf{30.65} & \textbf{10.33} & \textbf{23.96} & \textbf{21.65} \\
    \hline
  \end{tabular}
  }
  \caption{\label{tab:appendix_xsum}
    Comprehensive ROUGE scores on the XSum dataset. The scores for the text-only Token baseline are duplicated across all tree configuration columns to facilitate direct vertical comparison.
  }
\end{table*}
\end{document}